\documentclass{article}
\usepackage{ijcai23}

\usepackage{times}
\usepackage{soul}
\usepackage{url}
\usepackage[hidelinks]{hyperref}
\usepackage[utf8]{inputenc}
\usepackage[small]{caption}
\usepackage{graphicx}
\usepackage{amsmath}
\usepackage{amsthm}
\usepackage{bm}
\usepackage{amsfonts}
\usepackage{enumerate}
\usepackage{enumitem}
\usepackage{booktabs}
\usepackage{algorithm}
\usepackage{algorithmic}
\usepackage{multicol}
\usepackage{multirow}
\usepackage[switch]{lineno}
\usepackage{color}
\usepackage{colortbl}

\definecolor{grayy}{RGB}{230, 230, 230}
\definecolor{o}{RGB}{255, 140, 0}
\newcommand{\xf}[1]{{\color{black} #1}}
\newcommand{\gz}[1]{{\color{black} #1}}

\title{An Efficient Learning-based Solver Comparable to Metaheuristics \\ for the Capacitated Arc Routing Problem}

\author{
Runze Guo$^1$
\and
Feng Xue$^2$\and
Anlong Ming$^1$\And
Nicu Sebe$^2$
\affiliations
$^1$Beijing University of Posts and Telecommunications\\
$^2$University of Trento\\
\emails
grz@bupt.edu.cn,
feng.xue@unitn.it,
mal@bupt.edu.cn,
niculae.sebe@unitn.it
}

\begin{document}

\maketitle
 
\begin{abstract}
Recently, neural networks (NN) have made great strides in combinatorial optimization.
However, they face challenges when solving the capacitated arc routing problem (CARP) which is to find the minimum-cost tour covering all required edges on a graph,
while within capacity constraints.
In tackling CARP, NN-based approaches tend to lag behind advanced metaheuristics,
since they lack directed arc modeling and efficient learning methods tailored for complex CARP.
In this paper, 
we introduce an NN-based solver to significantly narrow the gap with advanced metaheuristics while exhibiting superior efficiency. 
First, we propose the direction-aware attention model (DaAM) to incorporate directionality into the embedding process,
facilitating more effective one-stage decision-making.
Second,
we design a supervised reinforcement learning scheme that involves supervised pre-training to establish a robust initial policy for subsequent reinforcement fine-tuning.
It proves particularly valuable for solving CARP that has a higher complexity than the node routing problems (NRPs).
Finally, a path optimization method is proposed to adjust the depot return positions within the path generated by DaAM.
Experiments illustrate that our approach surpasses heuristics and achieves decision quality comparable to state-of-the-art metaheuristics for the first time while maintaining superior efficiency.
\end{abstract}

\section{Introduction}

The capacitated arc routing problem (CARP) is a combinatorial optimization problem,
\xf{initially proposed by \cite{golden1981capacitated}.
This problem frequently arises in various domains such as inspection, monitoring, and search-rescue operations.}
\xf{The theoretical foundation of CARP is established on an undirected connected graph $\mathbf{G}=(\mathbf{V}, \mathbf{E}, \mathbf{E}_R)$,}
comprising a node set $\mathbf{V}$,
an edge set $\mathbf{E}$,
and a subset $\mathbf{E}_R \subseteq \mathbf{E}$ that needs to be served, called required edges.
\xf{Each required edge is associated with a specific demand value,
which is deducted from the remaining capacity of the vehicle upon working.
In this context, all vehicles start their routes from the depot node $depot \in \mathbf{V}$ and conclude their journey by returning to the same $depot$.
The primary objective of a CARP solver is to serve all the required edges with the lowest total path cost,
while adhering to the capacity constraint denoted as $Q$.}

\xf{According to \cite{golden1981capacitated}, the CARP is recognized as an NP-Hard problem,
making it impractical to obtain exact solutions for all CARP instances.
In the past,}
Memetic algorithms (MAs), classified as metaheuristic algorithms,
have achieved unrivaled results in addressing CARP challenges \cite{krasnogor2005tutorial,tang2009memetic}.
However, they \xf{have struggled} with high time costs and the exponential growth of the search space as the problem scale increases.
Compared to the traditional heuristics and metaheuristics,
NN-based solvers \cite{li2019learning,hong2022faster,RAMAMOORTHY2024200300} are faster with the assistance of GPU.
Thus, they have \xf{gained} increasing attention in recent years.
However, NN-based CARP solvers usually obtain paths with much lower quality compared to the traditional ones.
\xf{This discrepancy can be attributed to the following reasons:}
\begin{itemize}
\item
\textbf{Lack of edge direction in embedding learning:}
Existing methods model undirected edges instead of directed arcs,
\xf{which fails to encode edge directionality in embedding.}
As a consequence, they need to \xf{build edge sequences and determine edge directions separately}, 
leading to path generation without sufficient consideration.
\item
\textbf{Ineffective learning for solving CARP:}
\xf{CARP is more complex than Euclidean NRPs owing to the intricacies introduced by the non-Euclidean structure,
edge direction, and capacity constraints.
However, the advanced learning methods for NRPs are not directly transferable to solve CARP.
As a result,} there is a lack of effective learning schemes for tackling CARP.
\end{itemize}

In this paper,  
we aim to address both above issues and propose an NN-based solver for CARP that competes with the state-of-the-art MA \cite{tang2009memetic}.
Firstly,
we propose the direction-aware attention model (DaAM).
It computes embeddings for directed arcs rather than undirected edges,
thus avoiding missing direction information and enabling concise and efficient one-stage decision-making.
Secondly, we design a supervised reinforcement learning method to learn effective heuristics for solving CARP.
It pre-trains DaAM to learn an initial policy by minimizing the difference from the decisions made by experts.
Subsequently, DaAM is fine-tuned on larger-scale CARP instances by Proximal Policy Optimization with self-critical.
Finally,
to further boost the path quality,
we propose a path optimizer (PO) to re-decide the optimal return positions for vehicles through dynamic programming.
In the experiments,
our method demonstrates breakthrough performance,
closely approaching the state-of-the-art MA and surpassing classic heuristic methods with comparable efficiency.

\section{Related Work}
\subsection{Graph Embedding Learning}
Graph embedding \cite{cai2018comprehensive} aims to map nodes or edges in a graph to a low-dimensional vector space.
This process can be viewed as a learning process and is commonly achieved through popular graph neural networks (GNNs) \cite{wu2020comprehensive}.
Subsequently, GNN has derived many variants.
Kipf \textit{et al.} [\citeyear{kipf2016semi}] introduced graph convolutional operations to aggregate information from neighboring nodes for updating node representations.
\xf{Unlike} GCN,
GAT \cite{velivckovic2017graph} allowed dynamic node attention during information propagation \xf{by attention mechanisms}.
Other GNN variants \cite{hamilton2017inductive,wu2019simplifying} exhibited a similar information aggregation pattern but with different computational approaches.
\xf{In this paper, since an arc is related to the outgoing arc of its endpoint but irrelevant to the incoming arc of that,
we use attention-based methods to capture the intricate relationships between arcs for arc embedding learning.}


\subsection{Learning for Routing Problems}
The routing problem is one of the most classic combinatorial optimization problems (COPs),
and it is mainly categorized into two types according to the decision element:
node routing problems and arc routing problems.

\vspace{3pt}
\noindent
\textbf{Node routing problems} (NRPs),
such as the Traveling Salesman Problem (TSP) and Vehicle Routing Problem (VRP),
aim to determine the optimal paths traversing all nodes in either \gz{Euclidean space or graphs}.
As the solutions to these problems are context-dependent sequences of variable size,
they cannot be directly modeled by the Seq2Seq model~\cite{sutskever2014sequence}.
To address this problem,
Vinyals \textit{et al.} [\citeyear{vinyals2015pointer}] proposed the Pointer network (PN),
which achieves variable-size output dictionaries by neural attention,
and is applied to solving Euclidean TSP, Convex Hull, and Delaunay Triangulation.
Motivated by the scarcity of labels for supervised learning in COPs,
Bello \textit{et al.} [\citeyear{bello2016neural}] modeled the TSP as a single-step reinforcement learning problem and trained the PN using policy gradient \cite{williams1992simple} within Advantage Actor-Critic (A3C)~\cite{mnih2016asynchronous} framework. 
Nazari \textit{et al.} [\citeyear{nazari2018reinforcement}] observed the unordered nature of the input and replaced the LSTM encoder in PN with an element-wise projection layer.
Their model stood as the first NN-based approach employed to solve the Euclidean VRP and its variants.
To better extract correlations between inputs,
Kool \textit{et al.} [\citeyear{kool2018attention}] utilized multi-head attention for embedding learning.
They trained the model using REINFORCE \cite{williams1992simple} with a greedy baseline and exhibited outstanding results in experimental evaluations.
To solve COPs defined on graphs,
Khalil \textit{et al.} [\citeyear{khalil2017learning}] proposed S2V-DQN to learn heuristics for problems,
which employs structure2vec \cite{dai2016discriminative} for graph embedding learning and n-step DQN \cite{mnih2015human} for model training.
While the mentioned NN-based approaches have achieved comparable performance to metaheuristics,
they cannot be directly applied to solve ARP due to the modeling differences between ARP and NRP.

\vspace{3pt}
\noindent
\textbf{Arc routing problems} (ARPs) involve determining optimal paths for traversing arcs or edges in graphs.
Due to the complexity of the graph structure and directional constraints,
NN-based methods lag significantly behind traditional methods in solving ARPs.
\gz{Li and Li [\citeyear{li2019learning}] pioneered the use of the NN-based approach in solving the CARP by transforming it into an NRP.
They
first determined the sequence of edges and then decided the traversal direction for each edge.}
\xf{Hong and Liu [\citeyear{hong2022faster}] trained a PN in a supervised manner to select undirected edges in each time step, and also determined the edge traversal direction as post-processing.}
\gz{Ramamoorthy and Syrotiuk [\citeyear{RAMAMOORTHY2024200300}] proposed \xf{to} generate an initial tour based on edge embeddings and then split it into routes \xf{within} capacity constraint.}
\xf{These approaches} lack edge directionality encoding,
leading to edge selection without sufficient consideration and necessitating a two-stage decision process or an additional splitting procedure.
In contrast, our method directly models directed arcs,
enabling one-stage decision-making without extra process.

\section{Background}

The attention model (AM) \cite{kool2018attention} exhibits superior effectiveness in solving classic Euclidean COPs due to its attention mechanisms for extracting correlations between inputs.
Therefore, we use the AM as the backbone and give a brief review in terms of the TSP.

Given an Euclidean graph $\mathbf{G}\!\!=\!\!(\mathbf{V},\mathbf{E})$,
the AM defines a stochastic policy, denoted as $\pi(\bm{x}|\mathcal{S})$,
where $\bm{x}=(x_0,...,x_{|\mathbf{V}|-1})$ represents a permutation of the node indexes in $\mathbf{V}$,
and $\mathcal{S}$ is the problem instance expressing $\mathbf{G}$.
The AM is parameterized by $\bm{\theta}$ as:
\begin{gather}
\pi_\theta(\bm{x}|\mathcal{S})=\prod\nolimits_{t=1}^{|\mathbf{V}|} \pi_{\bm{\theta}} (x_t|\mathcal{S},\bm{x}_{0:t-1}),
\end{gather}
where $t$ denotes the time step.
Specifically,
the AM comprises an encoder and a decoder.
The encoder first computes initial $d_h$-dimensional embeddings for each node in $\mathbf{V}$ as $h^{0}_i$ through a learned linear projection.
It then captures the embeddings of $h^{0}_i$ using multiple attention layers, with each comprising a multi-head attention (MHA) sublayer and a node-wise feed-forward (FF) sublayer.
Both types of sublayers include a skip connection and batch normalization (BN).
Assuming that $l\in \{1,...,N\}$ denotes the attention layer,
the $l^{\text{th}}$ layer can be formulated as:
\begin{align}
\label{eq:am}
\hat{h_i} = & \ \text{BN}^l(h_i^{l-1} + \text{MHA}_i^l(h_0^{l-1}, \ldots, h_{|\mathbf{V}|-1}^{l-1})) \nonumber \\
h_i^l = & \ \text{BN}^l(\hat{h_i} + \text{FF}^l(\hat{h_i})).
\end{align}

The decoder aims to append a node to the sequence $\bm{x}$ \gz{at each time step}.
Specifically,
a context embedding $h_{(c)}$ is computed to \gz{represent the state at the time step $t$.}
Then a single attention head is used to calculate the probabilities for each node based on $h_{(c)}$:
\begin{align}
\label{eq:decode}
u_{(c)j} =&
\begin{cases}
C \cdot \tanh\left(\frac{[\textbf{W}^Qh_{(c)}]^T \textbf{W}^Kh_j^N}{\sqrt{d_h}}\right) & \text{if } j \ne x_{t^\prime} (\forall t^\prime < t) \\
-\infty & \text{otherwise},
\end{cases} \nonumber \\
p_i = & \ \pi_{\bm{\theta}} (x_t=i|\mathcal{S},\bm{x}_{0:t-1})= \frac{u_{(c)i}}{\sum\nolimits_j u_{(c)j}},
\end{align}
where $\textbf{W}^Q$ and $\textbf{W}^K$ are the learnable parameters of the last attention layer.
$u_{(c)j}$ is an unnormalized log probability with $(c)$ indicating the context node.
$C$ is a constant,
and $p_i$ is the probability distribution computed by the softmax function based on $u_{(c)j}$.

\begin{figure}
\centering
\includegraphics[width=0.48\textwidth]{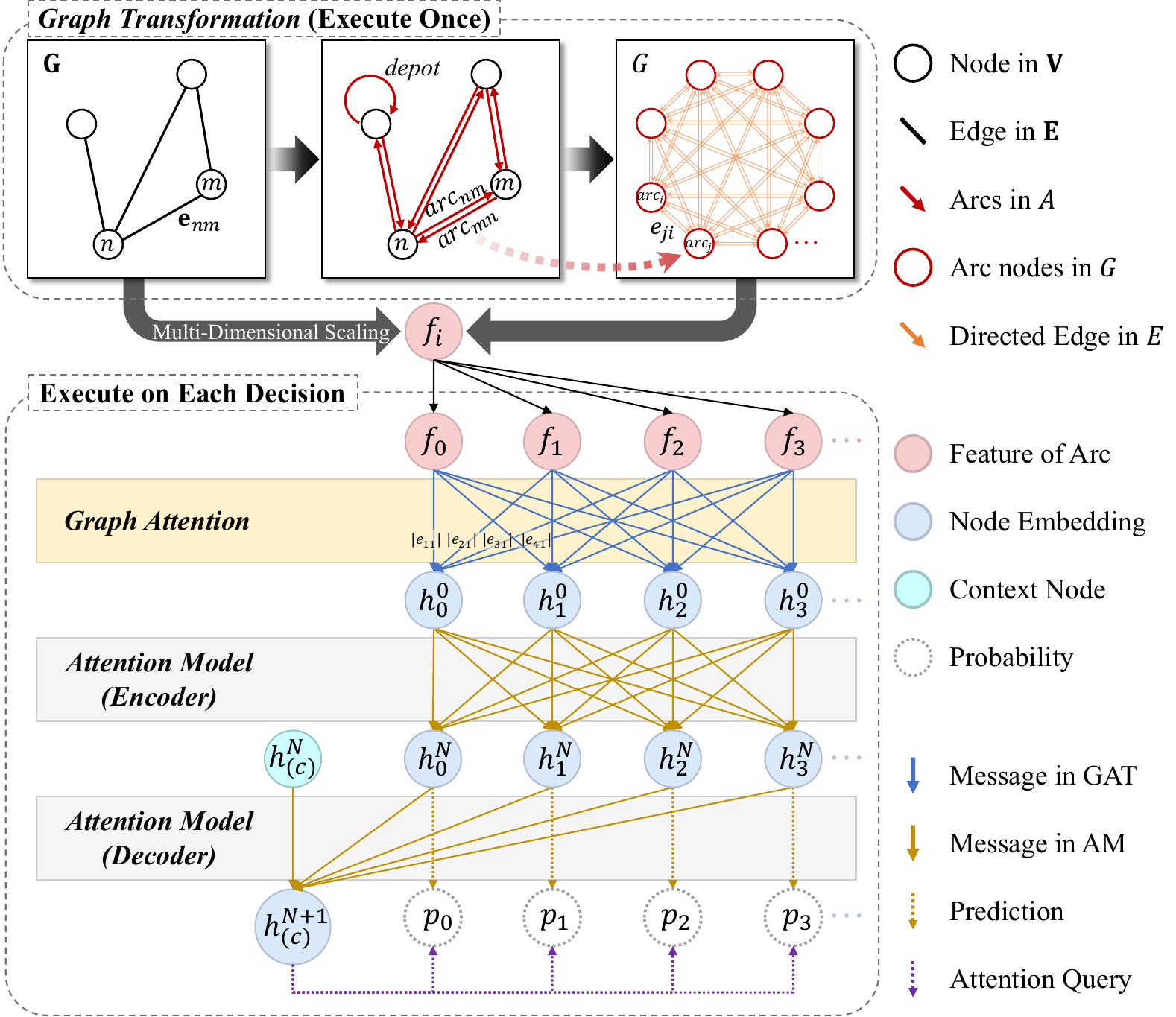}
\caption{\textbf{Pipeline of our DaAM} consists of two parts.
The first part transforms the input graph $\mathbf{G}$ by treating the arcs on $\mathbf{G}$ as nodes of a new directed graph $G$,
which only executes once in the entire pipeline.
The second part leverages the GAT and AM to update arc embeddings and select arcs,
which executes at each time step.}
\label{fig:pipeline}
\end{figure}

\section{Method}
\subsection{Direction-aware Attention Model}

In this section,
we propose the direction-aware attention model (DaAM).
Unlike previous methods that separately learn edge embeddings and determine edge directions,
our model encodes direction information directly into the embedding, enabling one-stage decision-making.

As shown in Fig. \ref{fig:pipeline},
the DaAM makes sequential decisions in two phases to select arcs.
\textbf{The first phase} is a one-time transformation process,
in which the arcs of the input graph are represented as nodes in the new directed complete graph.
\textbf{The second phase} is executed at each time step,
in which GAT is used to aggregate the weights between arcs.
Subsequently, AM is used to compute arc embeddings and generate the probability of selecting each arc.

\subsubsection{Arc Feature Formulation via Graph Transformation}
\vspace{3pt}
\textbf{Graph Transformation:}
Motivated by the need to consider direction when traversing edges,
we explicitly \gz{encode the edge direction} by edge-to-arc decomposition.
Let $\mathbf{G}\!\!=\!\!(\mathbf{V},\mathbf{E},\mathbf{E}_R)$ denotes the undirected connected graph as input,
where $\mathbf{V}$ is the node set of $\mathbf{G}$,
$\mathbf{E}$ is the edge set of $\mathbf{G}$,
and $\mathbf{E}_R\subseteq\mathbf{E}$ is the required edge set.
Firstly,
given that an edge has two potential traversal directions,
we decompose each edge $\mathbf{e}_{nm}\!=\!(cost_{nm},demand_{nm},allow\_serve_{nm}) \!\in\! \mathbf{E_R}$ into two arcs $\{arc_{nm}, arc_{mn}\}$ with opposite directions but the same cost, demand and serving state. 
\gz{Here $n, m$ are the indexes of node in $\mathbf{V}$.}
To simplify the representation below,
we replace $nm$ and $mn$ with single-word symbols, such as $i$ and $j$.
Edge decomposition results in a set of arcs denoted as $A_R$.
Secondly,
we \gz{build a new graph $G=(A_R,E)$}.
Specifically,
\gz{each arc in $A_R$ serves as a node in $G$,
and directed edge set $E$ is created,
with $e_{ij}\in E$ representing the edge from node $arc_i$ to $arc_j$.}
The weight $|e_{ij}|$ represents the total cost of the shortest path from the end node of $arc_i$ to the start node of $arc_j$.
In addition, 
we treat the depot as a self-loop zero-demand arc that allows for repeated serving,
denoted as $arc_0$.
Consequently,
we transform the input graph $\mathbf{G}$ into a directed complete \gz{graph $G$}.
\xf{By decomposing all edges in $\bm{E_R}$ into arcs,}
\gz{it is natural to directly select the arcs from $G$ during the decision-making, rather than the undirected edges.}

\begin{table}
  \centering
  \resizebox{\linewidth}{!}{
    \begin{tabular}{clll}
        \toprule
        No.  & Symbol & Field & Description\\
        \midrule
        1   & ${is\_depot}_i$                 & $\mathbb{F}_2$      & Does $arc_i$ correspond to the depot?\\
        2   & ${cost}_i$                      & $\mathbb{R}^+$      & Cost of $arc_i$. \\
        3   & ${demand}_i$                    & $\mathbb{R}^+$      & Demand of $arc_i$. \\
        4   & ${mds}_{\text{start}(i)}$       & ${\mathbb{R}^+}^d$  & Euclidean coordinates of $arc_i$'s start node.\\
        5   & ${mds}_{\text{end}(i)}$         & ${\mathbb{R}^+}^d$  & Euclidean coordinates of $arc_i$'s end node.\\
        6   & $|e_{x_{t\!-\!1} i}|$           & $\mathbb{R}^+$      & Edge weight from $arc_{x_{t\!-\!1}}$ to $arc_i$. \\
        7   & ${allow\_serve}_t^{(i)}$                & $\mathbb{F}_2$      & Is $arc_i$ at time step $t$ allowed to serve?\\
        \bottomrule
    \end{tabular}}
    \caption{\textbf{Feature Detail} of $arc_i$ at time step $t$ for CARP.}
    \label{tab:feature}
    \vspace{-5pt}
\end{table}

\vspace{3pt}
\noindent
\textbf{Arc Feature Formulation:}
To establish a foundation for decision-making regarding arc selection, 
the features of the arcs are constructed as input for the subsequent model. Specifically, 
multi-dimensional scaling (MDS) is used to project the input graph $\mathbf{G}$ into a $d$-dimensional Euclidean space. 
The Euclidean coordinates of $arc_i$'s start and end nodes, denoted as ${mds}_{\text{start}(i)}$ and ${mds}_{\text{end}(i)}$, 
are then taken as the features of $arc_i$ to indicate its direction.
As shown in Table \ref{tab:feature},
at time step $t$,
$arc_i$ can be featured as:
\begin{align}
    F_t^{(i)}=&({{is\_depot}}_i, {cost}_i, {demand}_i, {mds}_{\text{start}(i)}, {mds}_{\text{end}(i)}, \nonumber \\ 
    &|e_{x_{t\!-\!1}i}|,  {allow\_serve}^{(i)}_{t}), \quad t\in[1,+\infty),
\end{align}
where $x_{t\!-\!1}$ is the index of the selected arc at the last time step.

Our feature is modeled based on arcs rather than edges and encodes the direction attribute of arcs through MDS.
Therefore, compared to previous methods \cite{hong2022faster,li2019learning},
it is more suitable for ARPs that need to consider the direction of traversing edges.

\subsubsection{Arc Relation Encoding via Graph Attention Network}
Although AM is efficient in decision-making,
according to Eq. \eqref{eq:am},
it cannot encode the edge weights between nodes in $G$, an important context feature, during learning.
Therefore, 
we use graph attention network (GAT) \cite{velivckovic2017graph} to encode such weights.
At each time step $t$,
for each arc $arc_i$,
we integrate the weights between $arc_i$ and all arcs in $A_R$ along with their features into the initial embedding of $arc_i$.
\begin{align}
\label{eq:gat}
c_{ij}=&\ softmax\big(\alpha(\mathbf{W}[\,F_t^{(i)}\,||\,F_t^{(j)}\,|| \,|e_{ji}|\,])\big), \nonumber \\
h_i^0=&\ \sigma\big(\sum\nolimits_{j=0}^{|A_R|-1} c_{ij}\mathbf{W}F_t^{(j)}\big),
\end{align}
where $\mathbf{W}$ is a shared learnable parameter,
$[\cdot || \cdot]$ is the horizontal concatenation operator,
$\alpha(\cdot)$ is a mapping from the input to a scalar,
and $\sigma(\cdot)$ denotes the activation function.
$h_i^0$ denotes the initial feature embedding of $arc_i$,
which is taken as the input of subsequent AM.
\gz{Since $G$ is a complete graph, we use one graph attention layer to avoid over-smoothing~\cite{chen2020measuring}.}

\subsubsection{Arc Selection via Attention Model}
After aggregating the edge weights of $G$ into the initial embeddings,
we utilize AM to learn the final arc embeddings and make arc selection decisions.
In the encoding phase described by Eq.\ref{eq:am},
for each arc $\{arc_i\}$,
we leverage $N$ attention layers to process the initial embeddings $\{h_i^0\}$ and obtain the output embeddings of the $N^{\text{th}}$ layer,
i.e., $\{h_i^N\}$.
In the decoding phase,
we define the context node applicable to CARP:
\begin{gather}
h_{(c)}^N \!=\!
\Big[\frac{1}{|A_R|}\!\sum\nolimits_{i=0}^{|A_R|-1}\! h_i^N, h_{x_{t\!-\!1}}^N, \delta_t, \Delta_t\Big], t\in[1,+\infty)
\end{gather}
where $x_{t-1}$ indicates the chosen arc index at time step $t-1$ and $x_0$ is $arc_0$.
$\delta_t$ is the remaining capacity at time step $t$, $\Delta_t=\Delta(\delta_t > \frac{Q}{2})$ is a variable to indicate whether the vehicle's remaining capacity exceeds half.
Finally,
according to Eq.\eqref{eq:decode},
the decoder of AM takes the context node $h_{(c)}^N$ and arc embeddings $\{h_i^N\}$ as inputs and calculates the probabilities for all arcs, denoted as $p_i$.
The serviceable arc selected at time step $t$, i.e., $arc_{x_{t}}$, is determined by sampling or greedy decoding.

\subsection{Supervised Reinforcement Learning for CARP}
The decision-making of selecting arcs can be modeled as a Markov decision process with the following symbols regarding reinforcement learning:
\begin{itemize}
\item State $s_{t}$ is the newest path of arcs selected from $G$: $(arc_{x_0}, ..., arc_{x_{t-1}})$,
while the terminal state is $s_{T}$ with $T$ indicating the final time step.
\item Action $a_t$ is the selected arc at time step $t$, i.e., $arc_{x_t}$.
Selecting the action $a_t$ would add $arc_{x_t}$ to the end of the current path $s_{t}$ and tag the corresponding arcs of $arc_{x_t}$ with their features $allow\_serve$ changed to 0. Notably, $arc_0$ can be selected repeatedly but not consecutively.
\item Reward $r_t$ is obtained after taking action $a_t$ at state $s_t$, which equals the negative shortest path cost from the last arc $arc_{x_{t-1}}$ to the selected arc $arc_{x_t}$.
\item Stochastic policy $\pi(a_t|s_t)$ specifies the probability distribution over all actions at state $s_t$.
\end{itemize}
We parameterize the stochastic policy of DaAM with $\theta$:
\begin{gather}
\label{eq:model}
\pi (x_t|\ \mathcal{S},\bm{x}_{0:t-1})=\pi_{\theta} (a_t|s_t),
\end{gather}
where $\mathcal{S}$ is a CARP instance.
Starting from initial state $s_0$,
we get a trajectory $\tau=(s_0, a_0, r_0, ..., r_{T-1}, s_{T})$ using $\pi_{\theta}$.
The goal of learning is to maximize the cumulative reward: $R(\tau) = \sum\nolimits_{t=0}^{T-1} r_t$.
However, due to the high complexity of CARP,
vanilla deep reinforcement learning methods learn feasible strategies inefficiently.
A natural solution is to minimize the difference between the model's decisions and expert decisions.
To achieve this, we employ supervised learning to learn an initial policy based on labeled data and then fine-tune the model through reinforcement learning.

\subsubsection{Supervised Pre-training via Multi-class Classification}
In the pre-training stage, we consider arc-selection at each time step as a multi-class classification task,
and employ the state-of-the-art CARP method MAENS to obtain high-quality paths as the label.
Specifically,
assuming that $y_t\in\mathbb{R}^{|A_R|}$ denotes the one-hot label vector at time step $t$ of any path,
with $y_{t}^{(k)}$ indicating each element.
We utilize the cross-entropy loss to train the policy represented in Eq. \eqref{eq:model}:
\begin{gather}
L = -\sum\nolimits_{t=0}^{T-1} \sum\nolimits_{k=0}^{|A_R|-1} {y_{t}^{(k)} \log\big(\pi_\theta(arc_k|s_t)\big)}.
\end{gather}
We use the policy optimized by cross-entropy, denoted as $\pi_s$, to initialize the policy network $\pi_\theta$ and as the baseline policy $\pi_b$ in reinforcement learning.

\begin{algorithm}[t]
    \caption{PPO algorithm with self-critical}
    \label{alg:algorithm}
    \textbf{Input}: batch size $B$, number of episodes $K$, CARP train instance pool $\mathcal{P}$, CARP test instance pool $\mathcal{T}$ \\
    Initialize policies $\pi_\theta, \pi_{b} \leftarrow \pi_s$
    \begin{algorithmic}[1]
        \FOR{episode $k = 1$ to $K$}
            \STATE Initialize data batch $\mathcal{M}, \mathcal{M}^\prime \leftarrow ()$
            \WHILE{$|\mathcal{M}| < B$}
                \STATE Sample a CARP instance $\mathcal{S}$ from $\mathcal{P}$
                \STATE Sample $(s_0, a_0, s_1, a_1, \ldots, s_T)$ from $\mathcal{S}$ using $\pi_b$
                \STATE $\mathcal{M} \leftarrow \mathcal{M} \cup \{(s_0, a_0), (s_1, a_1), \ldots, (s_{T-1}, a_{T-1})\}$
            \ENDWHILE
            \FOR{each $(s, a) \in \mathcal{M}$}
                \STATE Sample trajectory $\tau_s^\theta$ using $\pi_\theta$ from $s$
                \STATE Greedily decode trajectory $\tau_s^b$ using $\pi_b$ from $s$
                \STATE Compute advantage $\mathcal{A}_s=R(\tau_s^\theta)-R(\tau_s^b)$
                \STATE $\mathcal{M}^\prime \leftarrow \mathcal{M}^\prime \cup \{(s, a, \mathcal{A}_s)\}$
            \ENDFOR
            \STATE Update $\pi_\theta$ using Adam over \eqref{eq:PPO} based on $\mathcal{M}^\prime$
            \IF{$\pi_\theta$ outperforms $\pi_b$ on $\mathcal{T}$}
                \STATE $\pi_b \leftarrow \pi_\theta$
            \ENDIF
        \ENDFOR
    \end{algorithmic}
\end{algorithm}

\subsubsection{Reinforcement Fine-tuning via PPO with self-critical}
During the fine-tuning phase,
we use Proximal Policy Optimization (PPO) to optimize our model $\pi_\theta(a_t|s_t)$ due to its outstanding stability in policy updates.
Considering the low sample efficiency in reinforcement learning, 
we employ a training approach similar to self-critical training \cite{rennie2017self} to reduce gradient variance and expedite convergence.
Specifically,
as shown in Algorithm \ref{alg:algorithm},
we use another policy $\pi_b$ to generate a trajectory and calculate its cumulative reward,
serving as a baseline function.
Our optimization objective is based on PPO-Clip~\cite{schulman2017proximal}:
\begin{align}
\label{eq:PPO}
&\mathbb{E}_{(s, a) \sim \pi_b} \Bigg[ \min\left(\frac{\pi_\theta(a|s)}{\pi_b(a|s)} \big(R(\tau_s^\theta)-R(\tau_s^b)\big), \right. \nonumber \\
&\left. \text{clip}\left(\frac{\pi_\theta(a|s)}{\pi_b(a|s)}, 1 - \epsilon, 1 + \epsilon \right) \big(R(\tau_s^\theta)-R(\tau_s^b)\big)\right)\Bigg],
\end{align}
where $s$ is used to replace current state $s_t$ for symbol simplification,
and $a$ for $a_t$.
$\text{clip}(w,v_{\min},v_{\max})$ denotes constraining $w$ within the range $[v_{\min},v_{\max}]$,
and $\epsilon$ is a hyper-parameter.
$\tau_s^\theta$ denotes a trajectory sampled by $\pi_\theta$ with $s$ as the initial state,
while $\tau_s^b$ for the trajectory greedily decoded by $\pi_b$.
In greedy decoding,
the action with the maximum probability is selected at each step.
$R(\tau_s^\theta)-R(\tau_s^b)$ serves as an advantage measure,
quantifying the advantage of the current policy $\pi_\theta$ compared to $\pi_b$.
We maximize Eq. \eqref{eq:PPO} through gradient descent,
which forces the model to select actions that yield higher advantages.
The baseline policy's parameters are updated if $\pi_\theta$ outperforms $\pi_b$.

\subsection{Path Optimization via Dynamic Programming}
The complexity of the problem is heightened by the increasing capacity constraint, 
making it challenging for the neural network to make accurate decisions regarding the depot return positions.
In this section,
we propose a dynamic programming (DP) based strategy to assist our model in optimizing these positions.

Assuming that $\bm{P}$ is assigned with the terminal state $s_T=(arc_{x_0},arc_{x_1},...,arc_{x_{T-1}})$,
representing a generated path.
Initially, we remove all the depot arcs in $\bm{P}$ to obtain a new path $\bm{P}^{'}=(arc_{x^\prime_0},arc_{x^\prime_1},...,arc_{x^\prime_{T^\prime-1}})$, where $\{x^\prime_i|i\!\in\![0,T^\prime\!-\!1]\}$ denotes a subsequence of $\{x_i|i\!\in\![0,T\!-\!1]\}$.
Subsequently,
we aim to insert several new depot arcs into the path $\bm{P}^{'}$ to achieve a lower cost while adhering to capacity constraints. 
To be specific,
we recursively find the return point that minimizes the overall increasing cost,
which is implemented by the state transition equation as follows:
\begin{align}
\label{eq:dp}
f(\bm{P}^{'}) = & \min_i (f(\bm{P}^{'}_{0:i}) +SC(arc_{x^\prime_i}, arc_{0}) \notag \\
& + SC(arc_{0}, arc_{x^\prime_{i+1}}) - SC(arc_{x^\prime_i}, arc_{x^\prime_{i+1}})), \notag \\
\text{s.t.} \quad 0 \leq & \ i < T^{'} - 1, \quad \sum\nolimits_{j=i+1}^{T^{'}-1} {demand}_{x^\prime_j} \leq Q, 
\end{align}
where $SC(arc_{x^\prime_i}, arc_0)\!=\!|e_{x^\prime_i 0}|$ denotes the shortest path cost from $arc_{x^\prime_i}$ to the depot.
$Q$ is the vehicle capacity.
According to Eq. \eqref{eq:dp},
we insert the depot arc $arc_0$ after an appropriate position $arc_{x^\prime_i}$,
which meets with the capacity constraint of the subpath $\bm{P}^{'}_{\smash{i+1:T^{'}-1}}$.
$f(\cdot)$ denotes a state featuring dynamic programming.
By enumerating the position $i$,
we compute the minimum increasing cost $f(\bm{P}^{'})$ utilizing its sub-state $f(\bm{P}^{'}_{0:i})$. 
The final minimum cost for path $\bm{P}$ is $f(\bm{P}^\prime) + g(\bm{P}^\prime)$, here $g(\bm{P}^\prime)$ is the unoptimized cost of $\bm{P}^\prime$.

We use beam search to generate two paths based on trained policy,
one under capacity-constrained and the other under unconstrained conditions. 
Both paths are optimized using DP and the one with the minimum cost is selected as the final result.

\begin{table}
  \centering
    \begin{tabular}{lcccc}
        \toprule
        CARP instances  & $|\mathbf{V}|$ & $|\mathbf{E}_R|$ &$demand$\\
        \midrule
        Task20      & 25-30     & 20 & 5-10\\
        Task30      & 30-35     & 30 & 5-10 \\
        Task40      & 45-50     & 40 & 5-10 \\
        Task50      & 55-60     & 50 & 5-10 \\
        Task60      & 65-70     & 60 & 5-10 \\
        Task80      & 85-90     & 80 & 5-10 \\
        Task100      & 105-110  & 100 & 5-10 \\
        \bottomrule
    \end{tabular}
    \caption{\textbf{Datasets information}.
    $|\mathbf{V}|$ is the number of nodes,
    $|\mathbf{E}_R|$ is the number of required edges.
    $demand$ represents the demand range for each required edge.
    Each dataset has 20,000 training instances and 10,000 test instances.}
    \label{tab:dataset}
\end{table}

\begin{table*}[htbp]
    \centering
    \resizebox{\textwidth}{!}{
        \begin{tabular}{@{}l|cc|cc|cc|cc|cc|cc|cc@{}}
            \toprule
            \multirow{2}{*}{Method} &
              \multicolumn{2}{c|}{Task20} &
              \multicolumn{2}{c|}{Task30} &
              \multicolumn{2}{c|}{Task40} &
              \multicolumn{2}{c|}{Task50} &
              \multicolumn{2}{c|}{Task60} &
              \multicolumn{2}{c|}{Task80} &
              \multicolumn{2}{c}{Task100} \\
             & Cost & Gap (\%) & Cost & Gap (\%) & Cost & Gap (\%) & Cost & Gap (\%) & Cost & Gap (\%) & Cost & Gap (\%) & Cost & Gap (\%) \\ \midrule
            \rowcolor{grayy} MAENS [\citeyear{tang2009memetic}]            & 474 & 0.00  & 706 & 0.00 & 950  & 0.00  &1222 & 0.00 & 1529 & 0.00  & 2113 & 0.00  & 2757 & 0.00  \\
            \midrule
            PS    [\citeyear{golden1983computational}]          & 544 & 14.72 & 859 & 21.76 & 1079 & 13.56 & 1448 & 18.45 & 1879 & 22.84 & 2504 & 18.49 & 3361 & 21.90 \\
            PS-Ellipse [\citeyear{santos2009improved}]   & 519 & 9.49 & 798 & 13.03 & 1006 & 5.89 & 1328 & 8.67 & 1709 & 11.77 & 2299 & 8.80 & 3095 & 12.26 \\
            PS-Efficiency [\citeyear{arakaki2019efficiency}] & 514 & 8.44 & 790 & 11.90 & 1007 & 6.00 & 1311 & 7.28 & 1684 & 10.14 & 2282 & 8.00 & 3056 & 10.85 \\
            PS-Alt1  [\citeyear{arakaki2019efficiency}]     & 514 & 8.44 & 791 & 12.04 & 1007 & 6.00 & 1312 & 7.36 & 1685 & 10.20 & 2283 & 8.04 & 3057 & 10.88 \\
            PS-Alt2  [\citeyear{arakaki2019efficiency}]     & 521 & 9.92 & 802 & 13.60 & 1009 & 6.21 & 1336 & 9.33 & 1720 & 12.49 & 2314 & 9.51 & 3102 & 12.51 \\
            S2V-DQN*  [\citeyear{khalil2017learning}]  & 590 & 24.42 & 880 & 24.65 & 1197 & 26.02 & 1520 & 24.32 & 1900 & 24.23 & 2820 & 33.43 & 3404 & 23.42 \\
            VRP-DL*  [\citeyear{nazari2018reinforcement}]  & 528 & 11.39 & 848 & 20.11 & 1193 & 25.57 & 1587 & 29.87 & 2033 & 32.96 & 2898 & 37.15 & 3867 & 40.26 \\
            \midrule
            DaAM (SL)  & 509 & 7.43 & 785 & 11.18 & 1066 & 12.24 &    -  & -     &  -    & -     &  -    & -  & - & -   \\
            DaAM (SL+RL) & 495 & 4.48  & 741 & 5.05 & 1009 & 6.19 & 1303 & 6.58 & 1639 & 7.16  & 2275 & 7.67  & 2980 & 8.06  \\
            DaAM (SL+RL+PO) &
              \textbf{482} &
              \textbf{1.65} &
              \textbf{725} &
              \textbf{2.73} &
              \textbf{992} &
              \textbf{4.39} &
              \textbf{1283} &
              \textbf{5.07} &
              \textbf{1621} &
              \textbf{5.98} &
              \textbf{2255} &
              \textbf{6.70} &
              \textbf{2958} &
              \textbf{7.28} \\ \bottomrule
        \end{tabular}
    }
    \caption{\textbf{Quantitative comparison}.
    All methods are evaluated on 10,000 CARP instances in each scale.
    We measure the gap (\%) between different methods and MAENS.
    Methods marked with an asterisk were originally proposed for NRP, but we modified them to solve CARP.
    The gray indicates that MAENS is taken as the baseline when calculating ``Gap''.
    Bold indicates the best results.}
    \label{tab:result}
  \vspace{-7pt}
\end{table*}

\section{Experiments}

\subsection{Experiment Setup}

\subsubsection{Problem Instances}

In our experiments,
we extracted the roadmap of Beijing, China randomly from OpenStreetMap \cite{haklay2008openstreetmap} to create CARP instances for both the training and testing phases.
All instances are divided into seven datasets, each representing different problem scales, as presented in Table \ref{tab:dataset}.
Each dataset consists of 30,000 instances, further divided into two disjoint subsets:
20,000 instances for training and the remaining for testing.
For each instance, the vehicle capacity is set to 100.

\subsubsection{Implementation Details}
Our neural network is implemented using the PyTorch framework and trained on a single NVIDIA RTX 3090 GPU.
The heuristics and metaheuristics algorithms are evaluated on an Intel Core i9-7920X with 24 cores and a CPU frequency of 4.4GHz.
We optimize the model using Adam optimizer \cite{kingma2014adam}.
The dimension of MDS coordinates $d$ is set to 8,
and the learning rate is set to $1e^{-4}$.
We set $\epsilon$ in the PPO training at 0.1.
Notably, our PPO training does not incorporate discounted cumulative rewards,
i.e., $\gamma$ is set to 1.

\subsubsection{Metrics and Settings}
For each method and dataset,
We compute the mean tour cost across all test instances, indicated by ``Cost''.
Employing the state-of-the-art MAENS \cite{tang2009memetic} as a baseline,
we measure the ``Cost'' gap between alternative algorithms and MAENS, indicated by ``Gap''.

For a more comprehensive comparison,
we compare our method against the heuristic Path-Scanning algorithms (PS) \cite{golden1983computational,santos2009improved,arakaki2019efficiency} and two NN-based algorithms.
Given the absence of publicly available code for prior NN-based CARP methods,
we modify two NN-based NRP solvers to suit CARP,
i.e, S2V-DQN \cite{khalil2017learning} and VRP-DL \cite{nazari2018reinforcement}.
Note that,
for S2V-DQN, we replace structure2vec with GAT to achieve more effective graph embedding learning.
For our method,
we incrementally add supervised pre-training (SL), reinforcement learning fine-tuning (RL), and path optimization (PO) to assess the effectiveness of our training scheme and optimization, respectively.

Due to the excessively long computation times of MAENS on larger-scale datasets,
SL is only performed on Task20, Task 30, and Task40.
The batch size for SL is set to 128. 
During the RL stage, greedy decoding is used to generate solutions,
and except for the Task20 dataset,
we utilize the training results obtained from the preceding smaller-scale dataset to initialize the model.
The beam width in the PO stage is set to 2.
For each dataset, we compare the mean cost of different methods on 10,000 problem instances. 

\begin{figure}[t]
\centering
\includegraphics[width=0.45\textwidth]{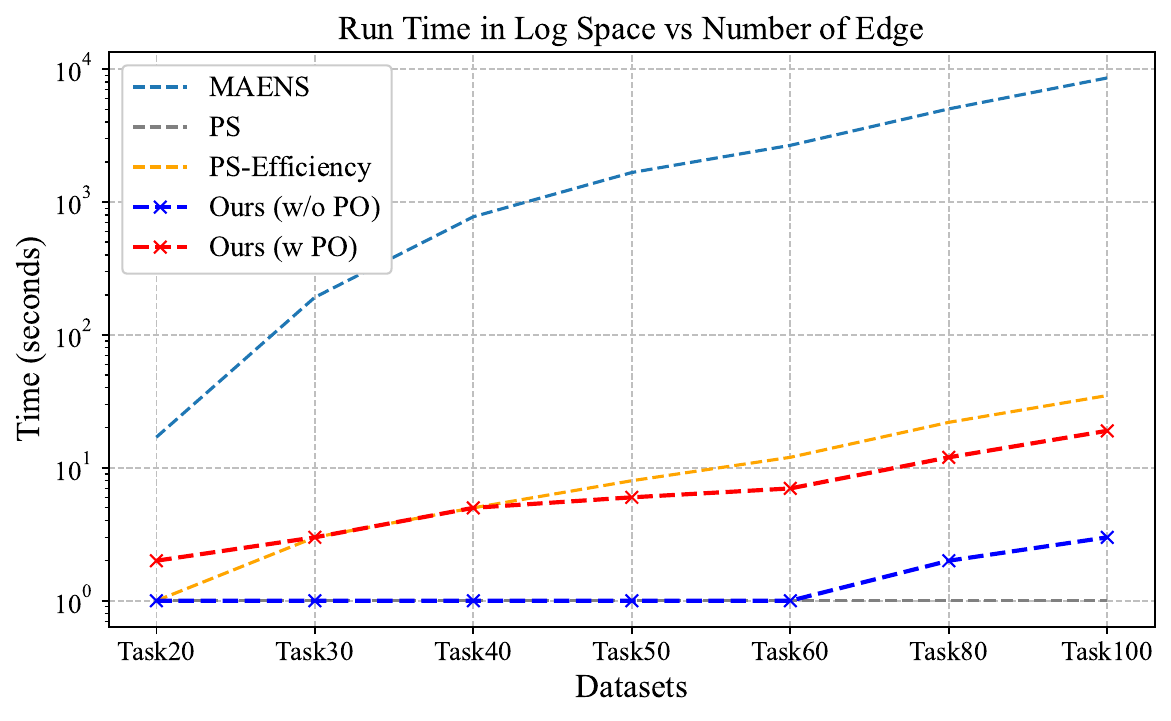}
\caption{\textbf{Comparison of run time}.
For each dataset, the mean time of each method on 100 CARP instances is shown.}
\label{fig:time}
  \vspace{-7pt}
\end{figure}
    
\subsection{Evaluation Results}

\subsubsection{Solution Quality}
Table \ref{tab:result} shows the result.
Our algorithm outperforms all heuristic and NN-based methods across all scales, achieving costs comparable to MAENS, trailing by less than $8\%$.
The advantage over PS demonstrates that neural networks can learn more effective policies than hand-crafted ones,
attributed to our well-designed modeling approach.
Moreover,
as the problem scale increases,
it becomes time-consuming to obtain CARP annotation by MAENS.
Therefore, we leverage the model pre-trained on small-scale instances as the initial policy for RL fine-tuning on Task50, Task60, Task80, and Task100,
yielding commendable performance.
This proves the generalization of our training scheme across varying problem scales.
The performance gap with MAENS highlights our algorithm's superiority in CARP-solving approaches.

\subsubsection{Run Time}
We compare the total time required for solving 100 CARP instances on all datasets using our method, MAENS, and PS algorithms,
and show the run time in log space.
For our method, we measured the solving time with and without PO.
Fig. \ref{fig:time} demonstrates that our method exhibits a significant speed advantage over MAENS,
even faster than PS-Efficiency \cite{arakaki2019efficiency} on Task50, Task60, Task80, and Task100.
In comparison, the consumption time of MAENS increases exponentially as the problem scale increases.
When dealing with large-scale CARP instances,
our method still generates paths efficiently with the aid of leveraging the data parallelism of GPUs and the instruction parallelism of CPUs.

\begin{figure}[]
  \centering
  \includegraphics[width=0.45\textwidth]{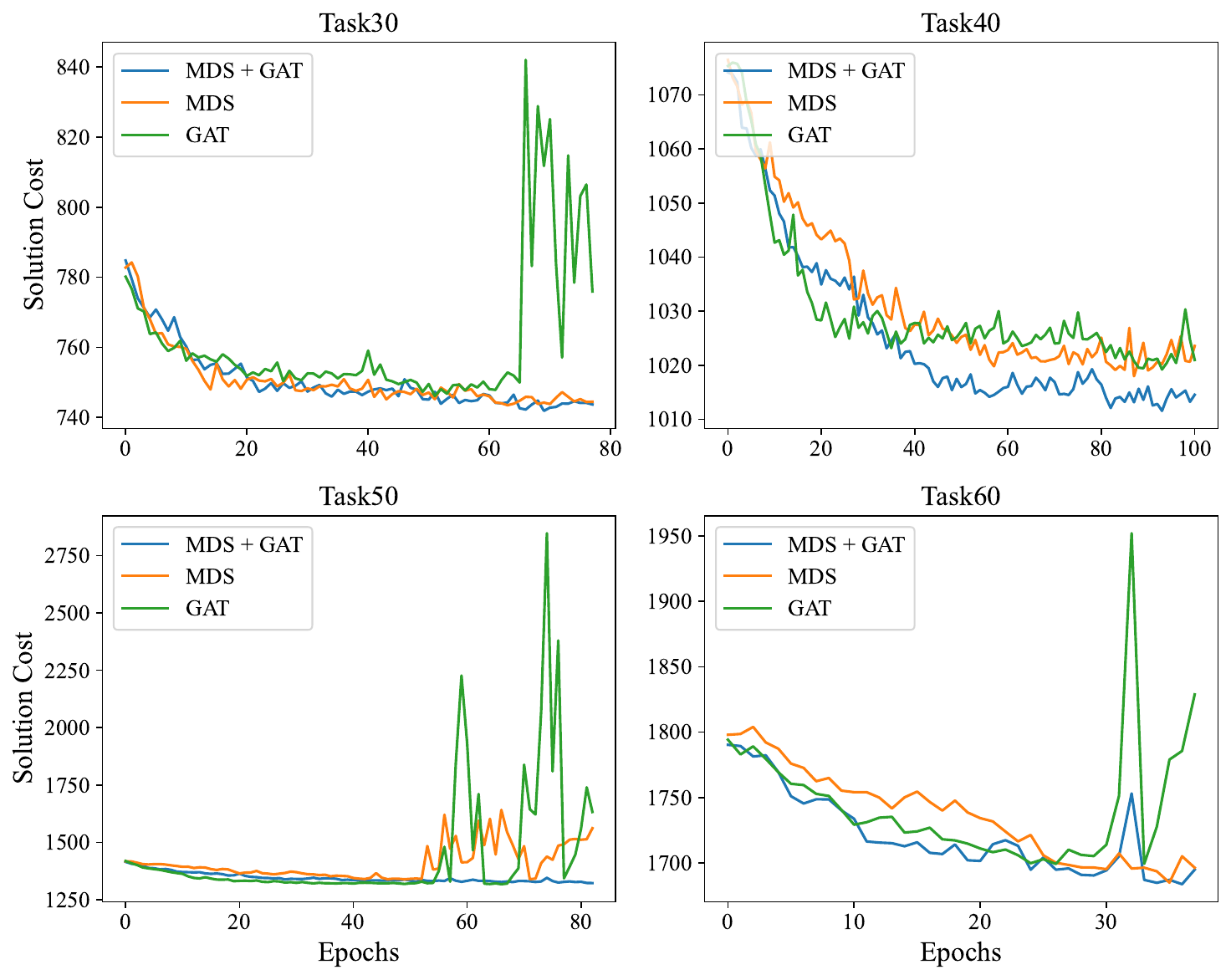}
  \caption{\textbf{Convergence trends} of different embedding learning methods in reinforcement learning training.}
  \label{fig:convergence}
  \vspace{-10pt}
\end{figure}

\begin{figure*}[t]
\centering
\includegraphics[width=1\textwidth]{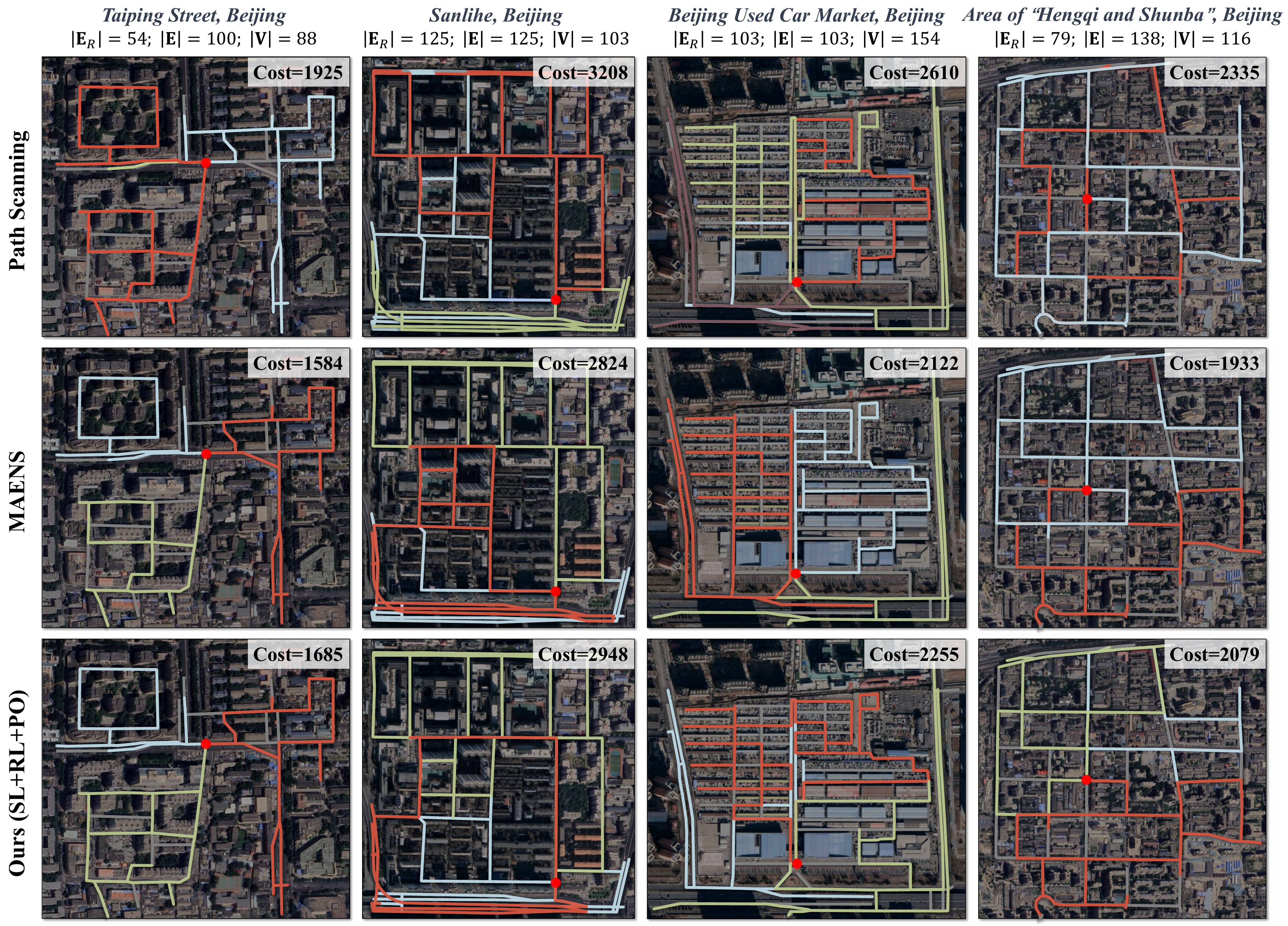}
\caption{\textbf{Qualitative comparison} in four real street scenes.
The paths are marked in different colors,
with gray indicating roads that do not require service and red points indicating depots.}
\label{fig:qualitative}
  \vspace{-5pt}
\end{figure*}

\subsubsection{Effectiveness of Combining MDS and GAT}
To evaluate the combination of MDS and GAT for embedding exhibiting,
we individually evaluate the performance of models using only MDS or GAT, as well as their combined performance.
The experiment is conducted on Task30, Task40, Task50, and Task60 by comparing the average performance of 1000 instances on each dataset.
In the RL stage, we use the policy pre-trained on Task30 for initialization.
Table \ref{tab:gat_mds} indicates that using MDS or GAT individually yields worse performance in most cases,
highlighting that combining MDS and GAT enhances the model's capacity to capture arc correlations.
Fig. \ref{fig:convergence} depicts the convergence trends in these scenes, which shows that the synergy between MDS and GAT contributes to the stability of training.

\subsubsection{Solution Visualization}

For a more intuitive understanding of the paths generated by different methods, 
we visualize and compare the results of our method with PS \cite{golden1983computational} and MAENS across four road scenes in Beijing.
Fig. \ref{fig:qualitative} visualizes all results alongside scene information.
We observe that our model obtains similar paths with MAENS
since we leverage the annotation generated by MAENS for supervised learning.
Furthermore, the paths generated by MAENS exhibit superior spatial locality,
clearly dividing the scene into distinct regions.
In contrast, the paths generated by PS appear more random.

\begin{table}[t]
\centering
\begin{tabular}{@{}l|c|c|c|c@{}}
\toprule
Method    & Task30       & Task40        & Task50        & Task60        \\ \midrule
MDS       & 743          & 1017          & 1338          & 1699          \\
GAT       & 746          & 1019          & \textbf{1317} & 1684          \\
MDS + GAT & \textbf{741} & \textbf{1011} & 1322          & \textbf{1683} \\ \bottomrule
\end{tabular}
\caption{\textbf{Costs of DaAM} using different embedding learning.}
\label{tab:gat_mds}
\vspace{-5pt}
\end{table}

\section{Conclusion}
In this paper, we propose a learning-based CARP solver that competes with state-of-the-art metaheuristics.
Firstly, we encode the potential serving direction of edges into embeddings,
ensuring that edge directionality is taken into account in decision-making.
Thus our method first achieves one-stage decision-making for ARPs.
Secondly, we present a supervised reinforcement learning approach that effectively learns policies to solve CARP.
With the aid of these contributions, our method surpasses all heuristics and achieves performance comparable to metaheuristics for the first time while maintaining excellent efficiency.

\appendix

\bibliographystyle{named}

\end{document}